%% file: main.tex
\documentclass[10pt,twocolumn]{article}

\usepackage{cvpr}
\usepackage{times}
\usepackage{epsfig}
\usepackage{graphicx}
\usepackage{amsmath}
\usepackage{amssymb}

\usepackage[utf8]{inputenc} % allow utf-8 input
\usepackage[T1]{fontenc}    % use 8-bit T1 fonts
\usepackage{url}            % simple URL typesetting
\usepackage{booktabs}       % professional-quality tables
\usepackage{amsfonts}       % blackboard math symbols
\usepackage{nicefrac}       % compact symbols for 1/2, etc.
\usepackage{microtype}      % microtypography
\usepackage{wrapfig}

\usepackage{multirow}
\usepackage[ruled]{algorithm2e}
\usepackage{sidecap}

\usepackage[moderate,mathdisplays=normal,wordspacing=normal]{savetrees}
\usepackage[usenames,dvipsnames]{xcolor}

\newcommand{\IGNORE}[1]{}

\include{kimkimacros}

\usepackage[pagebackref=true,breaklinks=true,letterpaper=true,colorlinks,bookmarks=false]{hyperref}

\cvprfinalcopy % *** Uncomment this line for the final submission

\def\cvprPaperID{3455} % *** Enter the CVPR Paper ID here

\ifcvprfinal\fi
\begin{document}

%%%%%%%%% TITLE
\title{Joint Manifold Diffusion for Combining Predictions on Decoupled Observations}

\author{Kwang In Kim\\
UNIST\\
\and
Hyung Jin Chang\\
University of Birmingham\\
}

\maketitle
%\thispagestyle{empty}

%%%%%%%%% ABSTRACT
\begin{abstract}
We present a new predictor combination algorithm that improves a given task predictor based on potentially relevant reference predictors. Existing approaches are limited in that, to discover the underlying task dependence, they either require known parametric forms of all predictors or access to a single fixed dataset on which all predictors are jointly evaluated. To overcome these limitations, we design a new non-parametric task dependence estimation procedure that automatically aligns evaluations of heterogeneous predictors across disjoint feature sets. Our algorithm is instantiated as a robust manifold diffusion process that jointly refines the estimated predictor alignments and the corresponding task dependence. We apply this algorithm to the relative attributes ranking problem and demonstrate that it not only broadens the application range of predictor combination approaches but also outperforms existing methods even when applied to classical predictor combination settings.
%\keywords{Predictor combination, unsupervised learning, diffusion}
\end{abstract}

\section{Introduction}
When the performance of an estimated predictor is not adequate for the task at hand, \eg due to limited training examples, we might benefit from the knowledge gained from related tasks. Multi-task learning (MTL)~\cite{ArgEvgPon08,KumDau12,LuoTaoGen13,PenShaLam15} explores this possibility by solving multiple problems simultaneously, and so capturing and benefiting from the potential task dependence. %The success of MTL in many visual learning problems has demonstrated their mutual dependence~\cite{ArgEvgPon08,KumDau12,YanRicSub14,PenShaLam15,KimTomRic17}.
The success of MTL in many visual learning problems has demonstrated such task dependence~\cite{ArgEvgPon08,KumDau12,YanRicSub14,PenShaLam15,KimTomRic17}. In most existing MTL algorithms, task dependence is modeled through the latent structures on the parameter spaces of the corresponding task predictors. %through the latent structures identified by the corresponding task predictors. Specifically, these algorithms introduces and enforces specific structures on the parameter spaces of task predictors. 
For instance, Evgeniou and Pontil's algorithm~\cite{EvgPon04} penalizes pair-wise parameter deviations of task predictors. Since it is unlikely that all tasks exhibit known task dependence structure, MTL algorithms attempt to automatically discover the underlying dependence and identify outliers, by \eg enforcing sparsity and/or introducing low-rank constraints on the aggregated task parameter matrices~\cite{ArgEvgPon08,GonYeZhang12}  %decomposing the task parameter matrix to submatrices each respectively specializes on discovering shared low-dimensional parameter spaces and identifying outliers~\cite{GonYeZhang12}, 
or by explicitly performing clustering of tasks~\cite{ZhoKwo12,PasRaiWai12}.

%When a single predictor is insufficient for a task, \eg due to limited training labels $L$, one could exploit auxiliary task information. For instance, transfer learning adapts a model trained on a different underlying probability distribution $Q(\mbx)$ into the current data generating distribution $P(\mbx)$. Multi-task learning (MTL) simultaneously trains predictors on multiple problems exploiting potential dependence present among tasks. These algorithms are limited in that they require access to the parametric forms and/or simultaneous training of of \emph{all} predictors of interest while often, in practical applications, the relevant predictors are available only as pre-trained models, pre-complied libraries, or even not available but instead indirectly given as predictions made on a data set $D'=\{\mbx_{n+1},\ldots,\mbx_m\}$. Kim~\etal proposed a framework that improves a given predictor $f$ of interest based on a set of potentially relevant predictors $H=\{h^k\}$. each obtained as a predictor of the respective learning problem. 

A major limitation of these traditional MTL approaches is that they require all task predictors to share the same predictive model or even the same parameter space, making them difficult to apply to heterogeneous predictors, \eg combining deep neural networks and support vector machines. However, the best predictor forms often depend on the individual tasks of interest. Further, existing MTL approaches are designed to train multiple predictors \emph{simultaneously}, and so they cannot be directly applied to train a new task predictor given previously-trained reference predictors, \eg combining pre-trained or pre-compiled predictor libraries without access to the corresponding task training data. 

Recently, Kim~\etal~\cite{KimTomRic17} proposed a non-parametric predictor combination approach where the predictor evaluations made at sampled data points are improved by combining them with reference predictions at \emph{test time} without requiring simultaneous training. This enables us to combine predictors with different or even unknown parametric forms. However, the application scope of this approach is limited in its own way, as it requires a large set of data points on which \emph{all predictors are jointly evaluated}. In practical applications, different predictions can be constructed based on the respective feature representations tailored for specific tasks of interest, and often these features are available by themselves as separate databases without having explicit references to the corresponding source data (\eg images). %Therefore, it is restrictive to require that all predictors are evaluated on a single fixed dataset.

In this paper, we propose a new algorithm to avoid the limitations of previous predictor combination approaches, thereby broadening the application spectrum of the non-parametric predictor combination approach~\cite{KimTomRic17}. Building on their test-time combination approach, our algorithm improves a task predictor based on a set of reference predictors. However, unlike their approach, we do not require that all predictors are available for evaluation on a single fixed set. Our algorithm takes as input \emph{decoupled} predictor evaluations %obtained from potentially heterogeneous features
and automatically aligns these predictions to discover the underlying task dependence. As the initial estimates of the alignments and the corresponding task dependence might be noisy, we denoise them jointly via a manifold diffusion process. The new algorithm combines the benefits of classical parametric MTL approaches and recent test-time combination algorithms, and facilitates combination applications where multiple heterogeneous predictors are constructed from disjoint feature sets. 
%Combining the benefits of classical parametric MTL approaches and recent test-time combination algorithms, our algorithm facilitates combination applications where multiple heterogeneous predictors are constructed from disjoint feature sets. 
We apply our algorithm to the \emph{relative attributes} ranking problem, and extend the application over previous approaches. Furthermore, evaluated on seven challenging datasets, our approach demonstrates that even when applied in the restricted settings of traditional approaches, it significantly improves both accuracy and time efficiency.

%\paragraph{Relative attributes ranking.} 
\textbf{Relative attributes ranking:} 
Relative attributes ranking~\cite{ParGra11,KovParGra12} refers to the problem of inferring a linear ordering of database images based on the strengths of attribute present in each entry. This problem differs from binary attribute classification where the goal is to predict the presence or absence of an attribute. Instead, relative attributes ranking focuses on attributes where such clear binary classifications cannot be obtained, \eg \textit{a shoe $A$ can be `more formal' than $B$ but it could still appear `less formal' than $C$.} %\eg a painting $A$ can be `brighter-in-color' than $B$ but it could be still appear `dimmer' than $C$.
This problem is also different from classical data-retrieval type ranking applications where the goal is to identify database entries that \emph{match a given query}.

This goal can be achieved by learning a rank function $f$ based on user-provided rank labels: Given a set of data points $X=\{\mbx_1,\ldots,\mbx_n\}\subset \calX$, rank learning aims to construct a function $f:\calX\to\R$ that agrees with the observed pair-wise rank labels $R=\{(i(1),j(1)),\ldots,(i(l),j(l))\}\subset X \times X$ where $(i,j)\in R$ implies that the rank of $\mbx_i$ is higher than $\mbx_j$: $f(\mbx_i)>f(\mbx_j)$. For instance, Parikh and Grauman's original \emph{Relative Attributes} algorithm learns a rank support vector machine (RankSVM)~\cite{ParGra11} while Yang~\etal extended it into \emph{Deep Relative Attributes}~\cite{YanZhaXu16} using neural networks.

%For both DNN and RSVM, we use the soft hinge loss: an ordered training pair $(i,j)\in R$ implies that the ranking of $\mbx_i$ should be higher than $\mbx_j$:
%\begin{align}
%l((\mbx_i,\mbx_j);f)=\max(0,1-f(\mbx_i)-f(\mbx_j))^2.
%\label{e:rankloss}
%\end{align}

%Rank learning was commonly used in data retrieval applications... . Recently, Parikh and Grauman proposed to learn attribute ranking functions that infers semantic attributes from database. This is an important problem by itself but it also facilities other applications, \eg zero-shot learning ... 
\section{Joint manifold diffusion for test-time predictor combination}
Our algorithm improves a given task predictor based on a set of reference predictors. As it is unknown a priori which reference predictors are relevant, our algorithm automatically identifies and exploits the relevant references. Existing approaches are limited in that they either require known and shared parametric forms for all task predictors (\eg in parametric MTLs) or evaluating multiple predictors on a single fixed dataset (in test-time predictor combination approach~\cite{KimTomRic17}). We bypass these limitations and allow the combination of multiple heterogeneous predictors by 1) a new non-parametric measure of task dependence (Sec.~\ref{s:predictormanifold}) and 2) a robust joint diffusion process that constructs bridge variables coupling the predictors of disjoint data instances (Sec.~\ref{s:jointdiffusion}).

\paragraph{Problem definition:}
Suppose that we are given a rank predictor function $f$ constructed as an estimate of the unknown ground-truth ranker (or task). % $t$.
Our goal is to \emph{refine} $f$ based on a set of $m$ \emph{reference} predictors $\{g^k\}_{k=1}^m$. As there is no guarantee that the reference predictors are relevant to the ground-truth task %$t$ 
or its estimate $f$, our algorithm automatically identifies any relevant references.

Adopting Kim~\etal's predictor combination framework~\cite{KimTomRic17}, we regard $f$ as a noisy observation of the ground-truth. %$t$. 
Our algorithm \emph{denoises} $f$ by embedding $\{f,g^k\}_{k=1}^m$ into a \emph{predictor manifold} $M$ and performing manifold denoising induced by the diffusion process therein. The domains of the predictor $f$ and references $\{g^k\}_{k=1}^m$ do not have to be identical. Instead, we assume that they are connected via an underlying data space $\widetilde{\calX}$ equipped with a probability distribution $P$. An example of $\widetilde{\calX}$ is the space of images while the corresponding data representation per task can be defined via the respective feature extractors $e^k:\widetilde{\calX}\to \calX^k$ on which the predictors are defined: $f\in C^\infty(\calX)$ and $g^k\in C^\infty(\calX^k)$.\footnote{Here, $C^\infty$ is the space of smooth (infinitely differentiable) functions.} Therefore, we regard a predictor $g^k$ being defined on its own feature domain $\calX^k$ or by combining it with the corresponding feature extractor, as a function on the shared data domain $\widetilde{\calX}$: $\tilde{g}^k:=g\cdot e^k\in C^\infty(\widetilde{\calX})$. As discussed shortly, this decomposition of feature representations and predictors facilitates applications where the main predictor $f$ is combined with multiple heterogeneous reference predictors.

%\paragraph{Predictor manifold.}
\subsection{Denoising over predictor manifold}
\label{s:predictormanifold}
%\noindent\textbf{Predictor manifold.}
%\noindent\textbf{Predictor manifold.} 
Assuming that the input space $\widetilde{\calX}$ is provided with a probability distribution $P$, our predictor manifold $M$ is given as an equivalence class of square-integrable functions $L^2(\widetilde{\calX},P)$: Each function $\tilde{g}\in L^2(\widetilde{\calX},P)$ is projected onto $M$ by centering and scale-normalization:
\vspace{-1mm}
\begin{align}
\label{e:manifoldprojection}
\text{Proj}_M[\tilde{g}]:=\frac{\tilde{g}-\int \tilde{g} dP }{\|\tilde{g}-\int \tilde{g} dP\|_{L^2(\widetilde{\calX},P)}}.
%\text{Proj}_M[\tilde{g}]:=\frac{\tilde{g}-\int \tilde{g}(\tilde{\mbx}) dP(\tilde{\mbx}) }{\|\tilde{g}-\int \tilde{g}(\tilde{\mbx}) dP(\tilde{\mbx})\|_{L^2(\widetilde{\calX},P)}}.
\end{align}
This manifold construction facilitates scale and shift-invariant comparisons of ranking functions: In ranking applications, \eg scaling of a ranker $g(\cdot)$ by a constant $c$, $cg(\cdot)$ should not alter the nature of rankings it induces. Similarly, a constant offset $g(\cdot)+c$ of a ranker $g(\cdot)$ should lead to the same ranking results. For problems where the absolute scales are important, \eg regression, inverse normalization can be performed after denoising. For brevity of notation, we omit the projection symbol $\text{Proj}_M$ and use $\tilde{g}$ to denote an element of $M$. The Riemannian\footnote{For $L^2(\widetilde{\calX},P)$, we adopt the natural identification of functions that deviates on a set of measure zero.} metric on this \emph{Hilbert sphere} $M$ can be induced from the ambient $L^2$ metric:
\begin{align}
%\langle \tilde{g}^k, \tilde{g}^l\rangle_{L^2(\widetilde{\calX},P)} = \int \tilde{g}^k(\tilde{\mbx})\tilde{g}^l(\tilde{\mbx})dP(\tilde{\mbx}),
\langle \tilde{g}^k, \tilde{g}^l\rangle_{L^2(\widetilde{\calX},P)} = \int \tilde{g}^k\tilde{g}^ldP,
\label{e:innerproduct}
\end{align}
which uniquely identifies a Laplace-Beltrami operator inducing a diffusion process on $M$.

It might be possible to evaluate the metric directly (Eq.~\ref{e:innerproduct}) if the parametric forms of the predictors $\{\tilde{f},\tilde{g}^k\}_{k=1}^m$ are known. When their parametric forms are unknown or for general non-parametric predictors, we instead approximate the metric $\langle \tilde{g}^k, \tilde{g}^l\rangle_{L^2(\widetilde{\calX},P)}$ based on their evaluations on a sample $\widetilde{X}=\{\tilde{\mbx}_1,\ldots,\tilde{\mbx}_n\}\subset \widetilde{\calX}$:
\begin{align}
\label{e:sampleevaluation}
\langle \mbf, \mbg^k \rangle =\frac{1}{n}(\mbf)^\top \mbg^k,\ \ \ \ \  \mbf:=\tilde{f}|_{\widetilde{X}},\ \ \ \mbg^k:=\tilde{g}^k|_{\widetilde{X}}.
\end{align}

%\kimki{Our manifold structure is inspired by Kim~\etal's predictor combination algorithm~\cite{KimTomRic17}: In~\cite{KimTomRic17}, each predictor is modeled as a Gaussian process (PG) with a mean function and covariance operator. Instead, we adopt the deterministic predictor model to simplify the technical exposition: A deterministic predictor can be regarded as a GP mean predictor while the covariance operator is fixed at Dirac delta function. }

\paragraph{Manifold denoising:} 
%\noindent\textbf{Manifold denoising.} 
%\noindent\textbf{Manifold denoising.} 
Using sample-based metric evaluations (Eq.~\ref{e:sampleevaluation}), the manifold denoising process can be described as to iteratively solve a diffusion equation on a graph formed by matrix $G=[\mbf^\top, (\mbg^1)^\top,\ldots,(\mbg^m)^\top]^\top\subset \R^{(m+1) \times n}$~\cite{WanTu13,HeiMai07,KimTomRic17}:  
\begin{align}
\frac{\partial G}{\partial t} = -\delta\Delta G
\label{e:diffusionequation}
\end{align}
with a diffusion coefficient $\delta>0$ and the graph Laplacian $\Delta$ constructed from $G$:
%where $\delta>0$ is a diffusion coefficient and $\Delta$ is the graph Laplacian constructed from $G$:
\begin{align}
\Delta &= I-D^{-\frac{1}{2}}W D^{-\frac{1}{2}},\nonumber\\
[W]_{kl} &= \exp\left(-\frac{\langle \mbg^k, \mbg^l \rangle^2}{\sigma^2}\right),
\label{e:gaussiankernel}
\end{align}
where $\sigma^2$ is a scale hyperparameter, and the diagonal matrix $D$ contains the row sums of $W$ ($[D]_{kk}=\sum_l W_{kl}$). When each graph node $\mbg^k$ corresponds to an i.i.d. Gaussian-noise contaminated observation of an underlying clean manifold point, this process tends to contract $G$ towards $M$~\cite{HeiMai07} and therefore, as the diffusion proceeds, $G(t)$ tends to recover a smooth noise-free version of $M$. 

To simulate the diffusion process, we discretize Eq.~\ref{e:diffusionequation} in time and obtain an implicit Euler update rule:
\begin{align}
\label{e:disceretediff}
G(t+1)-G(t) = -\delta \Delta(t) G(t+1).
\end{align}
Note the time-dependence of $\Delta$ as it is constructed from the variable 
$G$ being evolved (Eq.~\ref{e:gaussiankernel}).

\subsection{Joint manifold diffusion}
\label{s:jointdiffusion}
%\paragraph{$\mbf$-diffusion: Refining the predictor $\mbf$.} 
% \noindent\textbf{$\mbf$-diffusion: Refining the predictor $\mbf$.}
\subsubsection{$\mbf$-diffusion: Refining the predictor $\mbf$} 
%\noindent\textbf{$\mbf$-diffusion: Refining the predictor $\mbf$.} 
%At each time instance, the solution $G(t+1)$ can be obtained as the minimizer of a convex energy:
%\begin{align}
%\label{e:jointenergy}
%\calE(V) = \|G(t)-V\|^2_F+\delta \tr[V^\top \Delta V].
%\end{align}
As our goal is to refine the main predictor $\mbf$ given references, we hold the reference variables $\{\mbg^k\}_{k=1}^m$ in $G$ fixed and only update $\mbf$ during the diffusion (Eq.~\ref{e:disceretediff}). %: The updated solution $\mbf(t+1)$ at time $t+1$ (the first row of $G(t+1)$) can be obtained as the maximizer $\mbp^*$ of a score functional:\footnote{The implicit Euler step in Eq.~\ref{e:disceretediff} corresponds to a linear system whose solution can be obtained by minimizing the corresponding quadratic energy function; See~\cite{HeiMai07} for details.}
In this case, the updated solution $\mbf(t+1)$ at time $t+1$ (the first row of $G(t+1)$) can be obtained as the maximizer $\mbp^*$ of a score functional:\footnote{The implicit Euler step in Eq.~\ref{e:disceretediff} corresponds to a linear system whose solution can be obtained by minimizing the corresponding quadratic energy function; See~\cite{HeiMai07} for details.}
\begin{align}
\label{e:jointenergy}
\calO(\mbp) &= \langle\mbp, \mbf(t)\rangle_M^2+\delta \sum_{k=1}^{m} W_{1k} \langle \mbp, \mbg^k\rangle_M^2,
\end{align}
where we explicitly incorporate the normalization conditions (scaling and centering) such that the solution stays on the predictor manifold $M$:
\begin{align}
\langle \mba,\mbb\rangle_M=\frac{(C \mba)^\top C\mbb}{\|C \mba\|\|C \mbb\|}
%\langle \mba,\mbb\rangle_M=(C \mba)^\top C\mbb\big/\|C \mba\|\|C \mbb\|
\label{e:metric}
\end{align}
with $C=I-\frac{1}{n}\mbone \mbone^\top$ and $\mbone=[1,\ldots,1]^\top$. The score $\calO$ is a smooth function of $\mbp$, and it can be maximized using any smooth optimization method. However, by defining a symmetric matrix $Q=SS^\top$ with 
\begin{align}
\label{e:qmat}
S&=\left[\frac{C\mbf(t)}{\|\mbf(t)\|},\frac{\sqrt{\delta W_{11}} C\mbg^1}{\|C\mbg^1\|},\ldots,\frac{\sqrt{\delta W_{1m}} C\mbg^m}{\|C\mbg^m\|}\right],
\end{align}
%\begin{align}
%\label{e:qmat}
%Q&=\hat{\mbf}\hat{\mbf}^\top + \sum_{k=1}^m \hat{\mbg}^k(\hat{\mbg}^k)^\top,\\
%\hat{\mbf}&=\frac{C\mbf(t)}{\|\mbf(t)\|},\ \  \hat{\mbg}^k=\sqrt{\delta %W_{1k}}\frac{C\mbg^k}{\|C\mbg^k\|},
%\end{align}
it can also be rewritten as a generalized Rayleigh quotient
\vspace{-1mm}
\begin{align}
\label{e:jointenergyRayleigh}
\calO(\mbp) &= \frac{\mbp^\top Q \mbp}{\mbp^\top C \mbp}.
\end{align}
This reveals that the optimal solution $\mbp^*$ can be obtained as the eigenvector corresponding to the maximum eigenvalue of the generalized eigenvalue equation $Q\mbp=\lambda C\mbp$. For general symmetric matrices $Q$ and $C$, the computation for finding this eigenvector is cubic complexity: $O(n^3)$ for $n$ data points, which quickly becomes infeasible as $n$ grows. A more efficient approach can be taken by noting that for practical applications, the number of reference predictors $m$ will be much smaller than $n$ and the matrix $Q$ is constructed as a weighted combination of outer products of \emph{centered} vectors (Eq.~\ref{e:qmat}). Therefore, all eigenvectors $\{\mbe^k\}$ corresponding to non-zero eigenvalues of $Q$ are also centered, \ie, $\mbe^k=C\mbe^k$ implying that they also constitute the eigenvectors of the centering matrix $C$. This renders the generalized eigenvalue problem at hand into a regular eigenvalue problem $Q\mbp=\lambda \mbp$. Finally, the maximum eigenvector of $Q$ is obtained as the maximum left-singular vector of $S$ and hence the complexity of this step reduces to $O(m^2 n)$. As we maximize the squared metric in Eq.~\ref{e:jointenergy}, the optimizer $\mbp^*$ of $\calO$ can be inversely correlated to the original rank predictions $\mbf(0)$. Therefore, the final updated solution $\mbf(t+1)$ is obtained by multiplying the solution $\mbp^*$ with $\text{sgn}[-1\langle \mbp^*,\mbf(0)\rangle]$.

%\noindent\textbf{Discussion:} 
\paragraph{Discussion:} Our $\mbf$-diffusion step is motivated by \emph{adaptively-weighted} correction of $\mbf$ via \emph{robust local averaging} of the references $\{\mbg^k\}$. A key application challenge is that we do not know which references, if any, are relevant. Thus, our algorithm must \emph{automatically identify them}. This can be naturally addressed based on adaptive control of the combination weights $\{W_{1k}\}$ exercised via the diffusion process. Our algorithm controls the metric similarity between the main predictor and the references weighted by $\{[W]_{1k}\}$, which are increasing functions of the similarities themselves (Eqs.~\ref{e:jointenergy}-\ref{e:metric}). These weights provide the means to disregard irrelevant references. The uniformity of the weights is controlled by the hyperparameter $\sigma^2$ (Eq.~\ref{e:gaussiankernel}): For large $\sigma^2$, all references contribute equally, which might include outliers. For small $\sigma_w^2$, the single most relevant reference influences the solution, which might neglect other less relevant but still beneficial references. %Our algorithm automatically tunes $\sigma^2$ via validation. %Formally, our model manifold $M$ is the Hilbert sphere of functions with unit $L^2$ norm, embedded in the ambient $L^2(\widetilde{\calX},P)$ space. While the metric on $M$ is inherited from the inner-product structure of $L^2(\widetilde{\calX},P)$, we do not explicitly use it, the manifold denoising algorithm enables us to perform diffusion with only the ambient metric.

\subsubsection{$B$-diffusion: Combining predictions from decoupled observations}
%\noindent\textbf{$B$-diffusion: Combining predictions from decoupled observations.}
A major limitation of our initial predictor combination algorithm is that it relies on a large number of predictor evaluations sampled from the joint distribution $P(f,g^1,\ldots,g^m)$, \ie the sample predictions $\{\mbf,\mbg^k\}_{k=1}^m$ are obtained by \emph{jointly} evaluating the corresponding predictors $\{\tilde{f},\tilde{g}^k\}_{k=1}^m$ on a \emph{shared} sample set $\widetilde{X}\subset\widetilde{\calX}$. However, in practical applications, each predictor can be coupled with a feature representation tailored for an individual task of interest. Furthermore, often these features are available by themselves, without explicit references to the corresponding source images in $\widetilde{\calX}$. Therefore, even though the data generation processes of multiple feature domains $\{\calX,\calX^k\}_{k=1}^m$ are governed by a single probability distribution $P(\widetilde{\calX})$ on $\widetilde{\calX}$, it is unrealistic to assume that the available sample instances $\{X,X^k\}_{k=1}^m$ are all \emph{coupled}, \ie for all $i=\{1,\ldots,n\}$ and $k=\{1,\ldots,m\}$, there exists $\tilde{\mbx}_i\in \widetilde{\calX}$ such that  $\mbx^k_i=e^k(\tilde{\mbx}_i)\in X^k$. Also, the number of available sample instances may vary across tasks leading to predictor vectors that differ in sizes $\{\mbf,\mbg^k\}_{k=1}^m$. In this case, direct evaluations of the metric $\langle\cdot,\cdot\rangle_M$ in Eq.~\ref{e:jointenergy} is not possible.

Motivated by recent work on centered kernel alignment~\cite{RedBen16,CorMohRos12}, we construct \emph{bridge} variables $\{B^k\}_{k=1}^m$ that align each reference variable $\mbg^k$ to the main predictor variable $\mbf$. To motivate the construction, first we note that the metric evaluation $\langle\mbf,\mbg\rangle_M$ of prediction vectors $\mbf$ and $\mbg$ corresponds to a measure of the alignment of the corresponding centered gram matrices $G_\mbf=\mbf\mbf^\top$ and $G_\mbg=\mbg\mbg^\top$:
\begin{align}
\langle\mbf,\mbg\rangle_M=\frac{\tr[G_\mbf C G_\mbg C]}{\sqrt{\tr[G_\mbf C G_\mbf C]}\sqrt{\tr[G_\mbg C G_\mbg C]}}.
\label{e:centeredgramalign}
\end{align}

For typical kernel alignment applications \eg in kernel learning~\cite{CorMohRos12} and clustering~\cite{LuWanLu14}, a gram (kernel) matrix $G$ contains pair-wise evaluations of a positive definite kernel $k(\cdot,\cdot)$. In Eq.~\ref{e:centeredgramalign}, our kernel evaluates the product of two scalar inputs ($k(a,b)=ab$).

When the two gram matrices $G_\mbf$ and $G_\mbg$ are constructed from disjoint sample sets, and therefore, element-wise data coupling is not provided, a bridge matrix $B_{\mbg\mbf}$ of positive entries can be constructed to align $G_\mbg$ with respect to $G_\mbf$:
\begin{align}
\label{e:approxrho}
&\langle\mbf,\mbg\rangle_{B_{\mbg\mbf}}=\nonumber\\
&\frac{\tr[G_\mbf C B_{\mbg\mbf} G_\mbg B_{\mbg\mbf}^\top C]}{\sqrt{\tr[G_\mbf C G_\mbf C]}\sqrt{\tr[B_{\mbg\mbf} G_\mbg B_{\mbg\mbf}^\top C B_{\mbg\mbf} G_\mbg B_{\mbg\mbf}^\top C]}}.
\end{align}
The elements of each row in $B_{\mbg\mbf}$ total to one and therefore, each entry in the aligned gram matrix $B_{\mbg\mbf} G_\mbg B_{\mbg\mbf}^\top$ is obtained as a probabilistic (convex) combination of a $G_\mbg$-column.

If both gram matrices $G_\mbf$ and $G_\mbg$ are full rank as in existing kernel alignment applications, such a bridge matrix can be straightforwardly constructed by maximizing the alignment score $\langle\mbf,\mbg\rangle_{B_{\mbg\mbf}}$ (possibly, with additional regularizers, \eg non-negativity and sparsity~\cite{RedBen16}). Unfortunately, this approach is not applicable in our case as the number of variables in $B_{\mbg\mbf}$ is much higher than the effective degrees of freedom of the observed gram matrices (of rank 1): Our preliminary experiments indicated that na\"{i}vely applying this strategy trivially leads to the maximum alignment (value of 1), even for a random gram matrix $G_\mbg$.

Instead, we cast the bridge matrix learning as a continuous relaxation of bipartite graph matching: Suppose that $\mbf\in \R^{n(f)}$ and $\mbg^k\in\R^{n(k)}$ are obtained as evaluations of $f$ and $g^k$ on the respective feature instances  $X=\{\mbx_1,\ldots,\mbx_{n(f)}\}$, and $X^k=\{\mbx^k_1,\ldots,\mbx^k_{n(k)}\}$ and for each set, the first $n'$ data instances are paired, \ie there exists $\tilde{\mbx}_i\in \widetilde{\calX}$ such that  $([\mbf]_i,[\mbg^k]_i)=(f(\tilde{e}(\tilde{\mbx}_i)),g^k(\tilde{e}^k(\tilde{\mbx}_i)))$ for $i=1,\ldots,n'$. Using these coupling \emph{labels}, $B_{k\mbf}$ is initialized as
\begin{align}
[B_{k\mbf}(0)]_{ij} = \begin{cases}
    1 & \text{if } i=j \text{ and } i\leq n' \\
    0 & \text{otherwise.}
\end{cases}
\end{align}
which then evolves by diffusion propagating the labels to the entire bipartite graph $\calG=(X,X^k)$. To facilitate this process, we construct a pair of graph Laplacians $\Delta^\mbf$ and $\Delta^k$ based on the similarities of the respective feature domains and the predictor evaluations: For the main predictor $\mbf$, the Laplacian $\Delta^\mbf$ is defined as 
\begin{align}
\label{e:featurelaplacian}
\Delta^\mbf &= I-D^{-\frac{1}{2}}W^{\mbx\mbf} D^{-\frac{1}{2}},\ \ \ \ W^{\mbx\mbf}=W^\mbx_{ij}\circ W^\mbf_{ij},\\
W^\mbx_{ij} &= \exp\left(-\frac{\|\mbx_i-\mbx_j\|^2}{\sigma_\mbx^2}\right), W^\mbf_{ij} = \exp\left(\frac{([\mbf]_i-[\mbf]_j)^2}{\sigma_\mbf^2}\right),\nonumber
\end{align}
with $A\circ B$ being the Hadamard product of $A$ and $B$. The graph Laplacian $\Delta^k$ is similarly constructed. Note that $\Delta^\mbf$ and $\Delta^k$ are anisotropic as they use the corresponding predictor evaluations $\mbf$ and $\mbg^k$ in calculating the respective diffusivities ($W^{\mbx\mbf}$ and $W^{\mbx^k\mbg^k}$; Eq.~\ref{e:featurelaplacian}). Given the initial solution $B_{k\mbf}(0)$, the diffusion process on the bipartite graph $\calG$ is specified via these two Laplacians: The solution of the corresponding implicit Euler method is obtained as the minimizer of an energy
%\begin{align}
%\calE(V) &= \|V-B_{k\mbf}(0)\|^2_F+\delta_B(\tr[V^\top \Delta^\mbf V]+\tr[V \Delta^k V^\top])
%\label{e:implicitveuler}
%\end{align}
\begin{align}
\calE(V) &= \|V-B_{k\mbf}(0)\|^2_F\nonumber\\
&+\delta_B\tr[V^\top \Delta^\mbf V]+\delta_B\tr[V \Delta^k V^\top]
\label{e:implicitveuler}
\end{align}
whose optimum $V^*$ can be obtained as the solution of a Sylvester equation:
\begin{align}
\delta_B\Delta^\mbf V+\delta_BV \Delta^k=B_{k\mbf}(0).
\label{e:sylvester}
\end{align}
This analytical approach generates a dense matrix $B_{k\mbf}$, and therefore, it cannot be applied to large-scale problems ($n>10,000$). For these problems, we adopt the explicit Euler method and alternate $V$-updates based on two Laplacians:
\begin{subequations}
\begin{align}
\label{e:explicitveuler1}
B_{k\mbf}(t+1) &= B_{k\mbf}(t)-\delta_B \Delta^f B_{k\mbf}(t)\\
B_{k\mbf}(t+1) &= B_{k\mbf}(t)-\delta_B B_{k\mbf}(t)\Delta^k
\label{e:explicitveuler2}
\end{align}
\end{subequations}
explicitly controlling the sparsity of $B_{k\mbf}(t)$: At each iteration, each row of $B_{k\mbf}(t)$ is sparsified by keeping only the largest $K$ values and assigning zero to the rest of the elements. Given the initial label of $\{0,1\}$ in $B_{k\mbf}(0)$, the diffused variables $B_{k\mbf}$ stay bounded in $[0,1]$. At each iteration, we normalize each row of $B_{k\mbf}(t)$ such that its element values sum to 1.

% \paragraph{Joint diffusion.}
\subsubsection{Joint diffusion} 
%\noindent\textbf{Joint diffusion.} 
Our final algorithm consists of two diffusion processes: $\mbf$-diffusion updates the predictor variables $\mbf$ while $B$-diffusion updates the bridge variables. These diffusions are respectively governed by two classes of graph Laplacians $\Delta$ (Eq.~\ref{e:gaussiankernel}) and $\{\Delta^\mbf,\Delta^k\}_{k=1}^m$ (Eq.~\ref{e:featurelaplacian}), and as both $\Delta(t)$ and $\Delta^\mbf(t)$ depend on $\mbf(t)$, the two diffusion processes interact nonlinearly. We propose to interweave the two processes: First, we initialize $B$ by performing the $B$-diffusion. Then, the two steps of $\mbf$-diffusion and $B$-diffusion alternate until the termination condition is satisfied. Algorithm~\ref{a:mainalg} summarizes the proposed joint diffusion process.

\begin{algorithm}[t]
\caption{Predictor combination using joint manifold diffusion.}
\small{
\SetKwInput{Input}{Input}
\SetKwInput{Output}{Output}
\Input{Initial main predictor $\mbf$ and reference predictors $\{\mbg^k\}_{k=1}^m$; weight matrix $W^\mbx$ and reference graph Laplacians $\{\Delta^k\}_{k=1}^m$ (Eq.~\ref{e:featurelaplacian}); hyperparameters $\sigma^2$ (Eq.~\ref{e:gaussiankernel}), $\delta$ (Eq.~\ref{e:jointenergy}), $T_1$, and $T_2$;}
\Output{Refined predictions {$\mbf$}.}
\BlankLine
t = 0;\\
Build graph Laplacian $\Delta^{\mbx\mbf}$ using $W^\mbx$ and $\mbf(0)$ (Eq.~\ref{e:featurelaplacian});\\
\For{$t_1=1,\ldots,T_1$}{
\For{$t_2=1,\ldots,T_2$}{
Update $\mbf(t)$ based on the score function $\calO$ (Eq.~\ref{e:jointenergy}) and metric $\langle \cdot,\cdot\rangle_{B_{\mbg\mbf}}$ (Eq.~\ref{e:approxrho}).\\
$t = t+1$;\\
}
\For{$t_2=1,\ldots,T_2$}{
Update $\{B_{k\mbf}(t)\}_{k=1}^m$ based on Eqs.~\ref{e:implicitveuler}-\ref{e:explicitveuler2};\\
Normalize rows of $\{B_{k\mbf}(t)\}_{k=1}^m$;\\
$t = t+1$;\\
}
Update $\Delta^{\mbx\mbf}$ using $W^\mbx$ and $\mbf(t)$;\\
}
\label{a:mainalg}
}
\end{algorithm}

%\paragraph{Hyperparameters.} 
% \noindent\textbf{Hyperparameters.}
\subsubsection{Hyperparameters} 
Unlike the implicit Euler method (Eq.~\ref{e:implicitveuler}), the explicit $B_{k\mbf}$ update rule (Eqs.~\ref{e:explicitveuler1} and \ref{e:explicitveuler2}) is not stable uniformly over all values of $\delta_B$. Hence, we fix $\delta_B$ at a small value $10^{-5}$. Building the graph Laplacian $\Delta^\mbf$ (similarly for $\{\Delta^k\}_{k=1}^m$) requires tuning the scale parameters $\sigma_\mbx^2$ and $\sigma_\mbf^2$, and the number of nearest neighbors (NN) $N$ in $X$. We determine $\sigma_\mbx^2$ as twice the mean distance within the local $N$-neighborhood following Hein and Maier~\cite{HeiMai07}. The NN parameter $N$, the sparsity parameter $K$, and $\mbf$-scale parameter $\sigma_\mbf^2$ (similarly, $\sigma_k^2$) are globally tuned to maximize the maximum coupling score $\langle \mbf,\mbg^k\rangle_{B_{k\mbf}}$ across all reference $\{\mbg^k\}_{k=1}^m$ (Eq.~\ref{e:centeredgramalign}). They are determined during the first iteration and are held fixed throughout the diffusion process.

\begin{figure}
%\begin{SCfigure}
\centering
\includegraphics[width=0.65\linewidth]{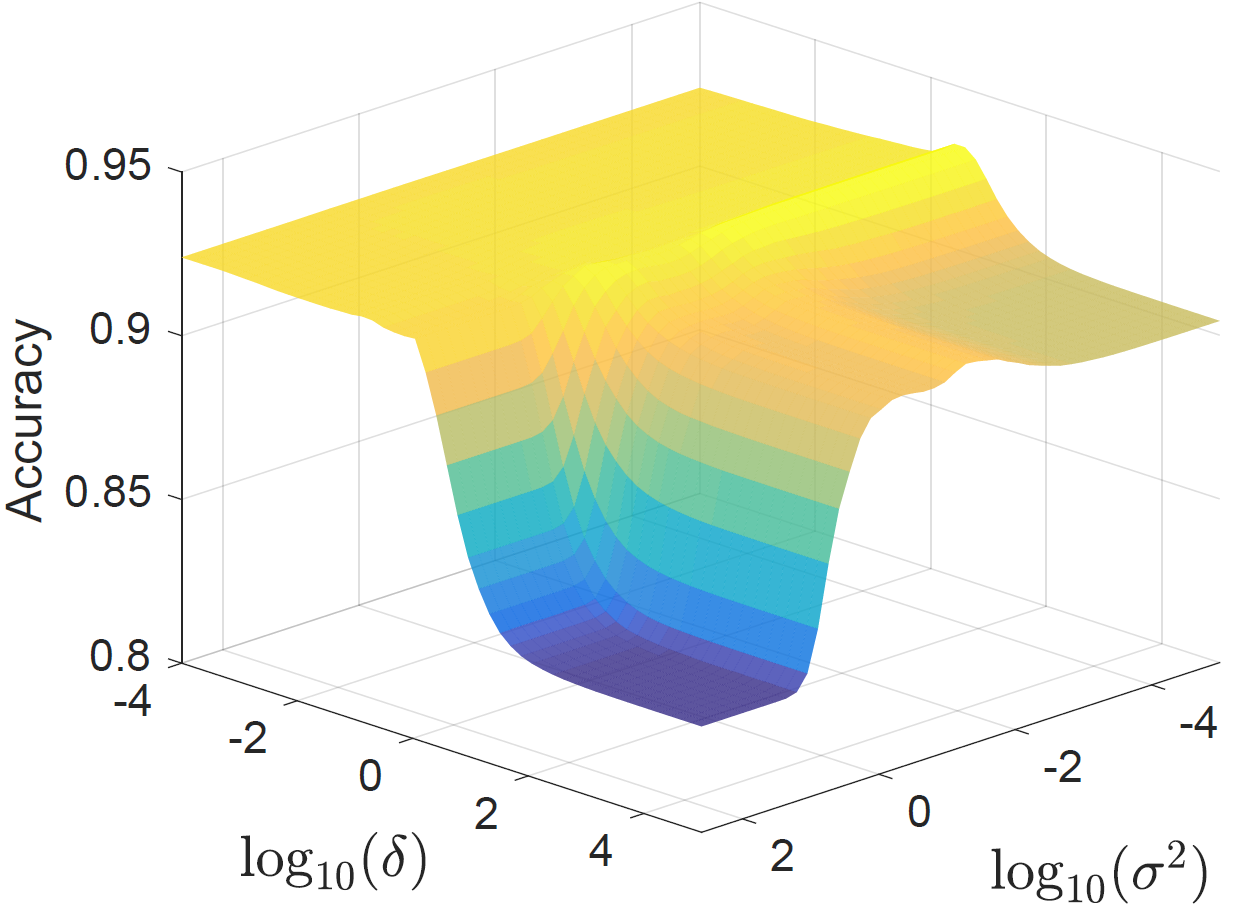}
%\caption{Accuracy of our algorithm on \textbf{OSR} dataset (attribute 3) with respect to hyperparameters $\sigma^2$ and $\delta$ varying over logarithmic intervals.}
\caption{Accuracy of our algorithm on \textbf{OSR} dataset (attribute 3) with respect to varying hyperparameters $\sigma^2$ and $\delta$.}
\label{fig:hyperparameters}
%\end{SCfigure}
\end{figure}

The step-size parameter $\delta$ (Eq.~\ref{e:jointenergy}) and the scale parameter $\sigma^2$ (Eq.~\ref{e:gaussiankernel}) for $\mbf$-diffusion is decided based on the ranking accuracy (defined as the ratio of correctly ranked pairs with respect to all pair-wise comparisons) on the validation sets: While our algorithm is unsupervised, we automatically tune the hyperparameters using small validation sets to facilitate fair comparisons with other algorithms (see Sec.~\ref{s:experiments} for details). In practice, the hyperparameters would be adjusted by the user trying different parameter combinations. Figure~\ref{fig:hyperparameters} shows that indeed, this sampling approach is feasible as the accuracy surface varies smoothly with respect to these hyperparameters. 

%The hyperparameters of our algorithm include the diffusion parameter $\delta$ (Eq.~\ref{e:jointenergy}) and the scale parameter $\sigma^2$ (Eq.~\ref{e:gaussiankernel}) which are all tuned based on the validation sets. 

For joint diffusion, we set an upper bound $T_2$ on the number of steps in each $\mbf$- and $B$-diffusion process, and terminate the iterations immediately when the validation accuracy (for $\mbf$-diffusion) or alignment score (for $B$-diffusion) does not increase. These two processes alternate until the joint iteration number meets the upper bound $T_1$, or the $\mbf$-validation accuracy does not improve. Our algorithm converges fairly quickly, typically within 10 iterations. We set $T_1,T_2=20$ (see Algorithm~\ref{a:mainalg}).

\section{Experiments}
\label{s:experiments}
\subsection{Design evaluation on a synthetic dataset}
%\noindent\textbf{Design evaluation on a synthetic dataset.}
To gain an insight into the effectiveness of our bridge estimation approach, we constructed a toy dataset with a known task metric structure. First, we generated 12 different tasks by explicitly building their ground-truth predictors $\{\tilde{t}^k\}_{k=1}^{12}$: Each member is constructed as a linear function on the $100$-dimensional input space: $\tilde{t}^k(\mbx) = \mbx^\top \widetilde{\mbw}^k$. Among the parameter vectors of 12 predictors, the last four are randomly generated (with each element sampled from the uniform distribution on $[-1,1]$) while the first 8 parameter vectors form two groups of 4 linearly depending predictors: $\widetilde{W}^1=[\widetilde{\mbw}^1,\ldots,\widetilde{\mbw}^4]$ is obtained by multiplying a pair of randomly generated vectors of sizes $100\times 1$ and $1\times 4$, respectively. The parameters of the second group (tasks 5-8) are generated similarly. The corresponding coupled noisy observations $H_\text{c}=\{\mbh_{\text{c}}^k\}_{k=1}^m$ are obtained by evaluating these ground-truths on an input dataset of $n$=1,000 data points $\widetilde{X}=\{\widetilde{\mbx}_1,\ldots,\widetilde{\mbx}_n\}$ and adding a mild level of noise (i.i.d. zero-mean Gaussian with standard deviation $0.2$) to the result. 

Similarly, decoupled observations $H_\text{d}=\{\mbh_\text{d}^k\}_{k=1}^m$ are constructed based on task-specific feature sets $\{X^k: X^k=\{\mbx^k_1,\ldots,\mbx^k_{n(k)}\}\}_{k=1}^m$ each sub-sampled from $\widetilde{X}$ ($n(k)\approx n/2$): To simulate different feature extraction operations, we applied principal component analysis with the feature dimensions varying randomly across tasks (under the condition that 95\% of the total variance is retained): $X^k\subset e^k|_{\widetilde{X}}$ with $e^k$ being the $k$-th principal component feature extractor. Finally, the noisy predictions $H_\text{d}$ are obtained by constructing the least-squares parameter approximations of $H_\text{c}$:
\begin{align}
\mbw^k = \argmin_{\mbw}\sum_{i=1}^{n(k)}(\mbw^\top \mbx_i^k-[\mbh_\text{c}^k]_i)^2,
\end{align}
evaluating the resulting predictors $\{g^k:g^k(\mbx)=\mbx^\top\mbw^k\}_{k=1}^{12}$ respectively on $\{X^k\}_{k=1}^{12}$, and adding Gaussian noise to the results. Across different feature matrices $\{X^k\}_{k=1}^{12}$, the source of feature instances in the first $30$ rows are shared, providing coupling labels. 

\begin{figure}[t]
\centering
\includegraphics[width=\columnwidth]{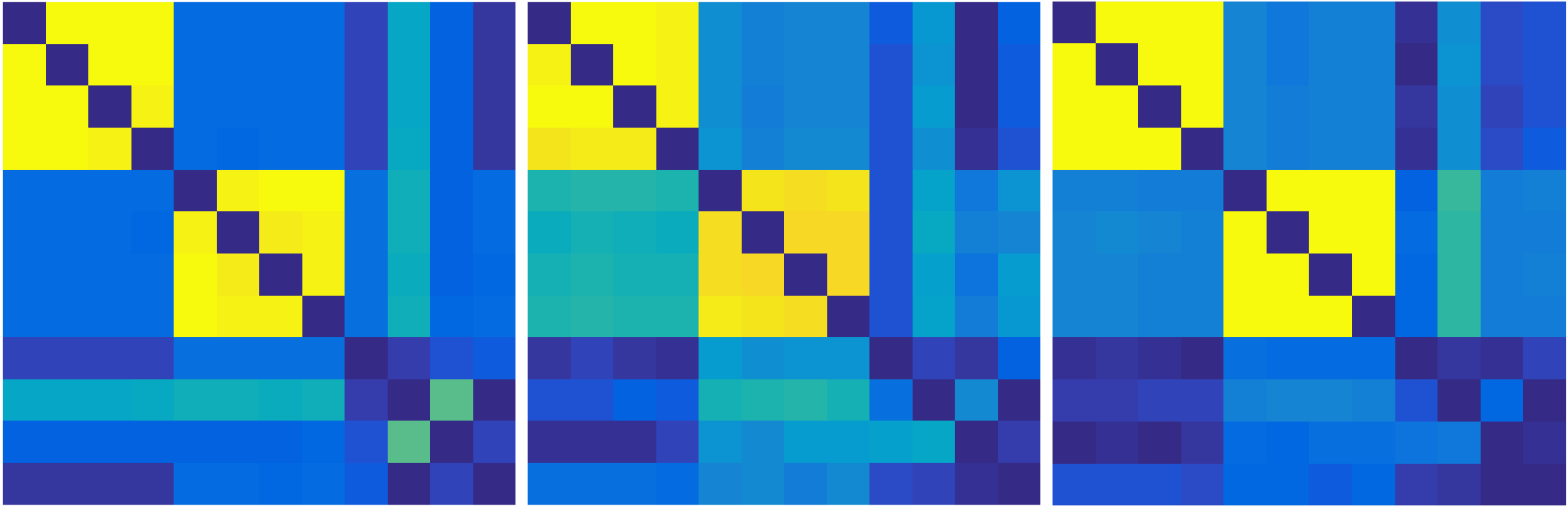}
\caption{Example estimation of the task metric $\langle\cdot,\cdot\rangle_M$ (Eq.~\ref{e:metric}) from decoupled predictions $\{\mbg^k\}_{k=1}^{12}$. By design, tasks 1-4 and tasks 5-8 respectively form groups of strongly correlated tasks. (left) pair-wise metric evaluations from the ground-truth predictions; (center) metric estimated based on decoupled predictions using the initial bridge estimate; (right) final metric evaluations constructed via joint diffusion.}
\label{f:taskmetric}
\vspace{-0.1cm}
\end{figure}

For each task $k$, we used $\mbh^k$ as the main predictor $\mbf$ and the rest as the references constituting a total of 12 predictor combination problems. Figure~\ref{f:taskmetric} shows the results of the bridge estimation process. (Left) shows the metric evaluated from the coupled predictions $H_\text{c}$: the $k$-row of the displayed matrix shows the metric evaluations of $\mbh_{\text{c}}^k$ (as the main predictor) with respect to the remaining predictors (as references). 
This matrix can be regarded as the ground-truths for bridge estimation process. (Center) shows the metric evaluated on the decoupled predictors $H_\text{d}$ using the initially estimated bridge variables $B_{k\mbf}(0)$ (Eq.~\ref{e:implicitveuler}). Given the mild level of task noise (as shown in Fig.~\ref{f:taskmetric}(left)), the initial metric evaluations on decoupled observations already well-recovered the underlying task dependence. Finally, (Right) shows the metric evaluated on the predictions denoised via the joint diffusion process. Our algorithm successfully suppressed noise and refined the underlying metric structure.

%The effectiveness of our algorithms may be improved in interactive scenarios, where users inspect different parameter combinations and chooses ones best-suited to their problem of interest.

%The number of diffusion updates are set at the maximum value of 50 and we immediate terminate the diffusion when the validation accuracy does not increase from the previous step. 

% \section{Experiments}
% \label{s:experiments}
\subsection{Evaluation on real datasets}
We evaluate our joint manifold diffusion algorithm on seven datasets and compare its performance with four baseline algorithms. Each entry in these datasets is assigned with multiple ground-truth attributes and therefore, predicting the relative strengths of these attributes constitutes multiple predictor combination problems: For each target attribute, our algorithm refines the corresponding predictor based on the remaining predictors as references.

% \paragraph{Experiment setting:}
\subsubsection{Baseline methods}
%\noindent\textbf{Setting.}
%\noindent\emph{Setting.}
\textbf{A) \emph{Ind}:}
The first baseline algorithm (\emph{Ind}) evaluates and selects the best predictor per dataset, per attribute from among deep neural networks (DNNs\cite{YanZhaXu16}), and linear and nonlinear rank support vector machines (RankSVMs\cite{ParGra11}) based on validation accuracy. For all experiments, the baseline algorithms were trained based on pair-wise rank labels extracted from 200 training data points. For given training inputs $X=\{\mbx_1,\ldots,\mbx_n\}$ and pair-wise rank labels $\{(i(1),j(1)),\ldots,(i(l),j(l))\}$, the linear RankSVM ($f(\mbx)=\mbw^\top \mbx$) minimizes the regularized rank energy:
\vspace{-2.5mm}
\begin{align}
\calE^S(f)=\sum_{k=1}^l L([\mbx_{i(k)},\mbx_{j(k)}],f) + \lambda^S \|\mbw\|^2,
\label{e:rankloss}
\end{align}
%\vspace{-3mm}
%where for each ordered pair $(\mbx_i,\mbx_j)$, the margin-based rank loss $L$ is defined as
where the margin-based rank loss $L$ is defined as
\begin{align}
L([\mbx_i,\mbx_j],f)=\left(\max(1-(f(\mbx_i)-f(\mbx_j)),0)\right)^2.
\end{align}
The regularization hyperparameter $\lambda^S\geq 0$ is tuned based on the accuracy on a separate validation set of the same size as the training set. For non-linear RankSVMs, we use a Gaussian kernel $k(\mbx,\mbx')=\exp\left(-\|\mbx-\mbx'\|^2/\sigma^2_S\right)$ 
%\begin{align}
%k(\mbx,\mbx')=\exp\left(-\frac{\|\mbx-\mbx'\|^2}{\sigma^2_S}\right)
%\label{e:svmkernel}
%\end{align}
with a scale hyperparameter $\sigma^2_S>0$. In this case, the parameter norm $\|\mbw\|^2$ in Eq.~\ref{e:rankloss} is replaced by the RKHS norm corresponding to $k$: $\|\mbw\|^2_k$.

\vspace{1.2mm}
\noindent\textbf{B) TPC:}
The second baseline uses Kim~\etal's test-time predictor combination approach (\emph{TPC})~\cite{KimTomRic17}. This algorithm was originally developed for regression but adapting it to ranking using rank loss $L$ is straightforward. Both \emph{TPC} and our algorithm require the initial main rank predictor $\mbf(0)$ and reference predictors $\{\mbg^k\}_{k=1}^m$ as inputs, which we obtain from \emph{Ind}. 

\noindent\textbf{C) \emph{MTL}$^1$:} The last two baselines (\emph{MTL}$^1$ and \emph{MTL}$^2$) implement adaptations of two existing multi-task learning algorithms. {\emph{MTL}$^1$} is based on Evgeniou and Pontil's approach of penalizing the pair-wise parameter deviations~\cite{EvgPon04}. Adapted to test time combination setting, \emph{MTL}$^1$ minimizes\footnote{Many other existing MTL approaches, \eg parameter matrix decomposition approaches~\cite{GonYeZhang12} and low-rank matrix learning algorithms~\cite{ArgEvgPon08}, strictly require simultaneous training, making them difficult to apply in the test-time combination setting of improving a predictor given \emph{fixed} references.} 
\vspace{-2.5mm}
\begin{align}
\calL_{\text{\emph{MTL}$^1$}}(f)&=\sum_{k=1}^l L([\mbx_{i(k)},\mbx_{j(k)}],f)\nonumber\\
 & + \lambda^S \|\mbw\|^2+\lambda^2 \sum_{k=1}^m W_k\|\mbw-\mbw^k\|^2
\label{e:mtl1}
\end{align}
where the weight parameters $\{W_k\}_{k=1}^m$ are defined similarly as in our task graph Laplacian (Eq.~\ref{e:gaussiankernel}): $W_k=\exp(-\|\mbw-\mbw^k\|^2/\sigma^2_\mbw)$.  %$W_k=\exp(-\frac{\|\mbw-\mbw^k\|^2}{\sigma^2_\mbw})$. 
The hyperparameters $\lambda^S$, $\lambda^2$, and $\sigma^2_\mbw$ are tuned based on a validation set.\footnote{Evgeniou and Pontil's original algorithm assumes that all tasks are related and therefore uses uniform weights, \ie $W_k=1/m$. Our preliminary experiments demonstrated that the non-uniform version (Eq.~\ref{e:mtl1}) always achieves higher accuracy indicating that not all tasks are equally relevant.} Similarly to RankSVM, \emph{MTL}$^1$ can also construct non-linear predictors using Gaussian kernels (with hyperparameter $\sigma^2_S$). 

\vspace{2mm}
\noindent\textbf{D) \emph{MTL}$^2$} adapts Pentina~\etal's curriculum learning approach~\cite{PenShaLam15}, which penalizes the deviation of the main predictor parameter $\mbw$ from a single best reference predictor $\mbw^k$. Pentina~\etal's original algorithm uses a bound on the generalization accuracy to select the reference predictor, which is not directly applicable to our rank learning problem. Instead, validation accuracy is used to select the reference.

For all datasets, we ran ten experiments with different training and validation set configurations and report the average results.

\vspace{-1mm}
%\noindent\emph{Datasets.} 
% \paragraph{Datasets.}
\subsubsection{Datasets}
%\noindent\textbf{Datasets.}
\textbf{A) Public Figure Face (PubFig)} dataset contains 800 images from 8 random identities~\cite{ParGra11}. Our goal is to estimate a linear ordering of database images based on the relative strengths of each of 11 different facial attributes (\emph{Masculine-looking}, \emph{White}, \emph{Young},  \emph{Smiling}, \emph{Chubby}, \emph{Visible-forehead}, \emph{Bushy-eyebrows} \emph{Narrow-eyes}, \emph{Pointy-nose}, \emph{Big-lips}, and \emph{Round-face}).

\vspace{1.2mm}
\noindent\textbf{B) Outdoor Scene Recognition (OSR)} dataset provides 2,688 images of 8 scene categories and 6 attributes~\cite{ParGra11}.

We use a combination of GIST features and color histograms for \textbf{PubFig} and GIST features for \textbf{OSR}. The attribute rank labels are constructed from the category labels as provided by the authors of \cite{ParGra11}. For each attribute, we improve the corresponding predictor using the predictors of the remaining attributes as references.

\vspace{1.2mm}
\noindent\textbf{C) Shoes} dataset contains 14,658 images of 10 categories and 10 attributes~\cite{KovParGra12}. We use a combination of GIST features and color histograms provided by the authors of~\cite{KovParGra12}. Our goal is to estimate the attribute rankings similarly to \textbf{PubFig} and \textbf{OSR} settings. However, here the datasets for the main and reference predictors are disjoint and we explicitly estimate the bride variables using additional 200 paired instances. As in this case \emph{TPC} is not applicable, we compare with \emph{MTL}$^1$ and \emph{MTL}$^2$.

\vspace{1.2mm}
\noindent\textbf{D) Cal7} dataset contains 1,474 images of 7 categories (\emph{Face}, \emph{Motorbikes}, \emph{Dolla-Bill}, \emph{Garfield}, \emph{Snoopy}, \emph{Stop-Sign}, and \emph{Windor-Chair}) as a subset of \textbf{Caltech-101} dataset~\cite{FeiFerPer07}. The dataset provides five different feature representations per image: wavelet, Gabor, CENTRIST, HOG, GIST, and LBP features~\cite{LiNieHua15}. The goal is to estimate a linear database ordering according to the category of each entry. For each single feature, we configured a corresponding main prediction task and constructed reference predictors using the remaining features. For each experiment, two disjoint feature sets for the main and reference predictors, respectively are prepared (roughly, half of the dataset was allocated for the main and the rest were allocated for references) representing the scenario where multiple predictions are generated based on heterogeneous, decoupled feature observations. To estimate the bridge variables, we use 200 coupled data instances as a sample from the joint distribution $P(f,g^1,\ldots,g^m)$. As the predictor variables are decoupled across tasks, \emph{TPC} is not applicable. Further, since the respective feature spaces and the corresponding predictors are heterogeneous, (adaptations of) classical parametric MTL approaches cannot be directly applied. Therefore, we compare our algorithm with only independent baselines (\emph{Ind}).

\vspace{1.2mm}
\noindent\textbf{E) NUS-WIDE-Object (NUS)} dataset contains 30,000 images of 31 categories~\cite{ChuTanHon09}. We use color histogram, color moments, color correlation, edge distribution and wavelet features as provided by the authors of \cite{ChuTanHon09} and \cite{LiNieHua15}.

\vspace{1.2mm}
\noindent\textbf{F) Handwritten digits (HW)} dataset provides 6 different feature representations of 2,000 handwritten digits, each represented by Fourier coefficients, profile correlations, Karhunen-Lo\`{e}ve coefficients, pixel averages in $2\times 3$ windows, Zernike moment and morphological features~\cite{BlaMer98}.

Experimental settings for \textbf{NUS} and \textbf{HW} are identical to \textbf{Cal7}. We use 200 paired data instances to learn bridge variables. 

\vspace{1.2mm}
\noindent\textbf{G) Animals With Attributes (AWA)} dataset contains 30,475 images of 50 animal categories. We use the SURF, SIFT and PHOG histograms and the features extracted by pre-trained DeCAF~\cite{DonJiaVin14} and VGG19~\cite{SimZis15} networks as provided by the authors of~\cite{LamNicHar09}. The experimental setting is similar to those of \textbf{Cal7}-\textbf{HW} except that, here we explicitly pair all data points across tasks enabling the application of \emph{TPC}. This \emph{toy setting} constitutes the ideal case where all reference predictors are inherently relevant in refining the main predictor and it enables us to verify the correct operation of \emph{TPC} and our approach.

\begin{figure*}[t]
\centering
\includegraphics[width=1.00\linewidth]{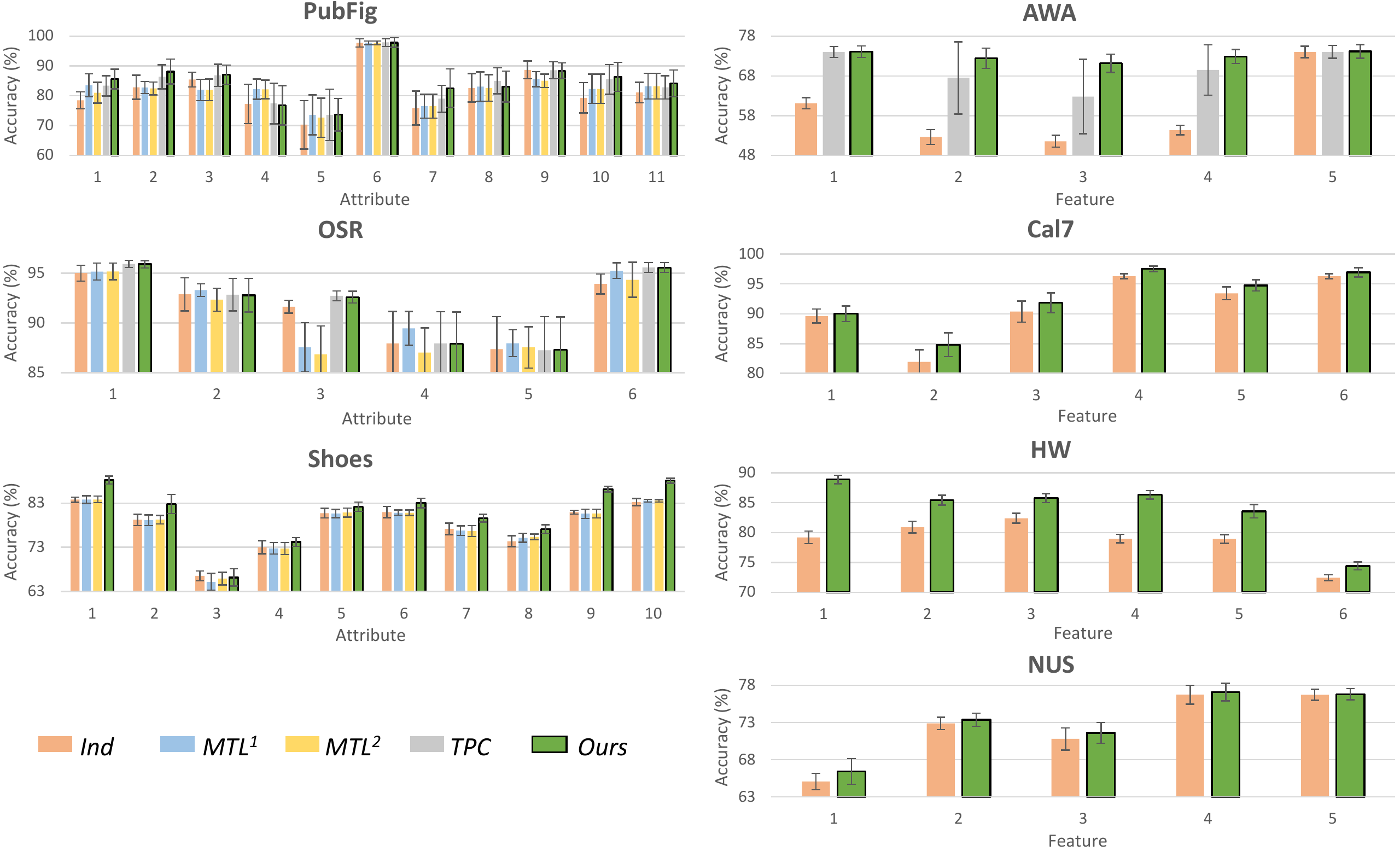}
\caption{Average accuracy of different ranking algorithms (over 10 different training and test set configurations. \emph{Ind}: best baseline independent predictors; \emph{MTL}$^1$ and \emph{MTL}$^2$: adaptations of existing MLT algorithms (\cite{EvgPon04} and \cite{PenShaLam15}, respectively); \emph{TPC}: Kim~\etal's test-time predictor combination algorithm~\cite{KimTomRic17}. The length of each error bar corresponds to twice the standard deviation.}
\label{f:rankresults}
% \vspace{-0.2cm}
\end{figure*}

\vspace{-1mm}
% \paragraph{Results.}
\subsubsection{Results}
%\noindent\textbf{Results.}
%\noindent\emph{Results.} 
%\kimki{Table~\ref{t:rankresults}}?
Figure~\ref{f:rankresults} summarizes the results. While not all target attributes show marked improvements, \emph{TPC} and our algorithm consistently improve upon or are on par with \emph{Ind}. Comparing \emph{TPC} and ours, the performances are almost identical on \textbf{OSR}. For \textbf{PubFig}, the two algorithms demonstrated the complementary strengths across different target attributes, while our algorithm achieves higher average accuracy. The corresponding results on \textbf{AWA} are notably different: While \emph{TPC} already achieves better results than the baseline \emph{Ind}, our algorithm further improves accuracy by a large margin. In addition, by virtue of the fast Eigen-decomposition-based approach (Eq.~\ref{e:jointenergyRayleigh}) the runtime of our algorithm is around 20\emph{$\times$} shorter than TPC: For \textbf{AWA} with 30,475 images, our algorithm took around 0.2 seconds for the entire combination process. As \emph{TPC} requires fully coupled predictor evaluations, it cannot be applied to \textbf{Cal7}, \textbf{NUS}, and \textbf{HW} datasets, in which our algorithm continues to outperform \emph{Ind}. For these datasets, our algorithm demonstrates even better performance than the best individual task predictors, which demonstrates the utility of combining predictors across multiple features.

The two multi-task learning adaptations \emph{MTL}$^1$ and \emph{MTL}$^2$ to the test-time combination setting also showed measurable performance improvement over \emph{Ind}. In particular, they achieved the highest average accuracy on target attributes 2, 4, and 5 of the \textbf{OSR} dataset. On the other hand, for \textbf{PubFig} and \textbf{Shoes}, our algorithm constantly outperformed these algorithms demonstrating complementary strengths. As both \emph{MTL}$^1$ and \emph{MTL}$^2$ require the parametric forms of all predictors to be shared across different tasks, it is not straightforward to apply these algorithms when different tasks use heterogeneous features (\textbf{Cal7}, \textbf{NUS}, and \textbf{HW} datasets).

%\kimki{Parameter sensitivity}
%Here we consider the problem where different data sets of the same underlying problems are provided in terms of multiple feature attributes.

\section{Conclusion}
In this paper, we have presented a new algorithm improving a given task predictor by combining multiple reference predictors, each constructed from the respective tasks. 
Conventional approaches require either all task predictor's known and shared parametric forms or multiple predictors' evaluation on a single fixed dataset. 
We address these limitations by formulating the problem as a non-parametric task dependence estimation and by a robust joint diffusion process that automatically couples the predictors of disjoint data instances. 
This not only facilitates a new (decoupled, parameter-free) predictor combination application but also significantly improves the accuracy and run-time over existing algorithms when applied to challenging relative attributes ranking datasets.

Our manifold structure (Eq.~\ref{e:manifoldprojection}) and metric therein (Eqs.~\ref{e:innerproduct}--\ref{e:sampleevaluation}) are directly aligned with the case when the predictor outputs are one-dimensional (\eg ranking and regression problems). When the output space is multi-dimensional (\eg multi-class classification), our metric structure needs changing to align predictions of different dimensions. We expect that this can be done by calculating canonical correlations between the input pairs, but it would involve non-trivial modifications.

% \kimki{This part could be removed if necessary:} 
Identifying data coupling across heterogeneous domains is a challenging problem. This problem arises in the predictor combination setting where different predictors are evaluated on data instances sampled from multiple heterogeneous domains. We attempted to address this challenge by estimating \emph{soft} couplings via a joint diffusion process propagating a small set of coupled data points. An alternative possibility that we have not explored in this work is to consider recent label-free set pairing approaches, \eg instantiated using cyclic GANs~\cite{ZhuParIso17}. This type of approach is not immediately applicable to our setting as they do not generate explicit pairings and, therefore, would require modifying the entire task dependence measure and the corresponding denoising process. Future work should explore this possibility.

{\small
\bibliographystyle{ieee_fullname}
\bibliography{biblio}
}

\end{document}

%% file: kimkimacros.tex
\usepackage{mathtools}
\usepackage{xspace}

\newcommand{\mbone}{\mathbf{1}}
\newcommand{\mba}{\mathbf{a}}
\newcommand{\mbb}{\mathbf{b}}

\newcommand{\mbe}{\mathbf{e}}
\newcommand{\mbf}{\mathbf{f}}
\newcommand{\mbg}{\mathbf{g}}
\newcommand{\mbh}{\mathbf{h}}

\newcommand{\mbp}{\mathbf{p}}

\newcommand{\mbw}{\mathbf{w}}
\newcommand{\mbx}{\mathbf{x}}

\newcommand{\R}{\mathbb{R}}

\newcommand{\calE}{\mathcal{E}}

\newcommand{\calG}{\mathcal{G}}

\newcommand{\calL}{\mathcal{L}}

\newcommand{\calO}{\mathcal{O}}

\newcommand{\calX}{\mathcal{X}}

\def\R{\mathbb{R}}

\def\calX{\mathcal{X}}

\def\tr{\mathop{\rm tr}\nolimits}

\def\argmin{\mathop{\rm arg\,min}\limits}%    a math operator.

\newcommand{\ignore}[1]{}

\makeatletter
\DeclareRobustCommand\onedot{\futurelet\@let@token\@onedot}
\def\@onedot{\ifx\@let@token.\else.\null\fi\xspace}
\def\eg{{e.g}\onedot} %\def\Eg{{E.g}\onedot}
\def\ie{{i.e}\onedot} %\def\Ie{{I.e}\onedot}

\def\etal{{et al}\onedot}
\makeatother